\documentclass[letterpaper, 10 pt, conference]{ieeeconf}  

\IEEEoverridecommandlockouts                              

\usepackage{graphics} 
\usepackage{epsfig} 
\usepackage{mathptmx} 
\usepackage{times} 
\usepackage{amsmath} 
\usepackage{amssymb}  
\usepackage{amsfonts}

\usepackage{algorithm}
\usepackage{algpseudocode}
\usepackage{booktabs} 
\usepackage{tabularx}  
\usepackage{float}

\usepackage{tikz}
\usetikzlibrary{arrows.meta, positioning, shapes.geometric, shapes.multipart}
\usetikzlibrary{shapes.geometric, arrows.meta, positioning, fit, backgrounds}

\title{\LARGE \bf
Distilling Reinforcement Learning Policies for Interpretable Robot Locomotion: Gradient Boosting Machines and Symbolic Regression
}

\author{Fernando Acero and Zhibin Li
\thanks{F. Acero is supported by the EPSRC CDT in Foundational AI [EP/S021566/1] at the UCL Centre for Artificial Intelligence.}
\thanks{Department of Computer Science, University College London
         {\tt\small \{fernando.acero, alex.li\}@ucl.ac.uk}}%
}

\begin{document}


\maketitle
\thispagestyle{empty}
\pagestyle{empty}

\begin{abstract}

Recent advancements in reinforcement learning (RL) have led to remarkable achievements in robot locomotion capabilities. However, the complexity and ``black-box'' nature of neural network-based RL policies hinder their interpretability and broader acceptance, particularly in applications demanding high levels of safety and reliability. This paper introduces a novel approach to distill neural RL policies into more interpretable forms using Gradient Boosting Machines (GBMs), Explainable Boosting Machines (EBMs) and Symbolic Regression. By leveraging the inherent interpretability of generalized additive models, decision trees, and analytical expressions, we transform opaque neural network policies into more transparent ``glass-box'' models. We train expert neural network policies using RL and subsequently distill them into (i) GBMs, (ii) EBMs, and (iii) symbolic policies. To address the inherent distribution shift challenge of behavioral cloning, we propose to use the Dataset Aggregation (DAgger) algorithm with a curriculum of episode-dependent alternation of actions between expert and distilled policies, to enable efficient distillation of feedback control policies. We evaluate our approach on various robot locomotion gaits -- walking, trotting, bounding, and pacing -- and study the importance of different observations in joint actions for distilled policies using various methods. We train neural expert policies for 205 hours of simulated experience and distill interpretable policies with \textbf{only 10 minutes} of simulated interaction for each gait using the proposed method.  

\end{abstract}

\section{INTRODUCTION}

Explainability and interpretability are topics of increasing relevance in artificial intelligence and robotics \cite{gunning2019xai, sakai2022explainable, milani2023explainable}. Whilst reinforcement learning (RL) has enabled significant advancements in robot locomotion over model-based optimization \cite{lee2020learning, yang2020multi, miki2022learning, defazio2024learning}, existing work has ubiquitously used neural networks for representing policy and value functions due to their general function approximation capabilities and automatic gradient-based optimization, making them suitable for policy gradient algorithms widely used in RL. 

However, as robots transition out of research environments into industrial or domestic applications where they can deliver value to society, the black-box nature of neural networks ushers significant challenges in terms of interpretability and explainability, arguably rendering them unsuitable for safety-critical or consumer-facing use cases that particularly require behaviour or system certification \cite{milani2023explainable}. We note that many interpretable models such as decision trees or symbolic expressions do not easily allow for generic gradient-based optimization. Because of this, there is a motivation to transform neural locomotion policies into interpretable ones. 



\begin{figure}
    \centering
    \includegraphics[width=\columnwidth]{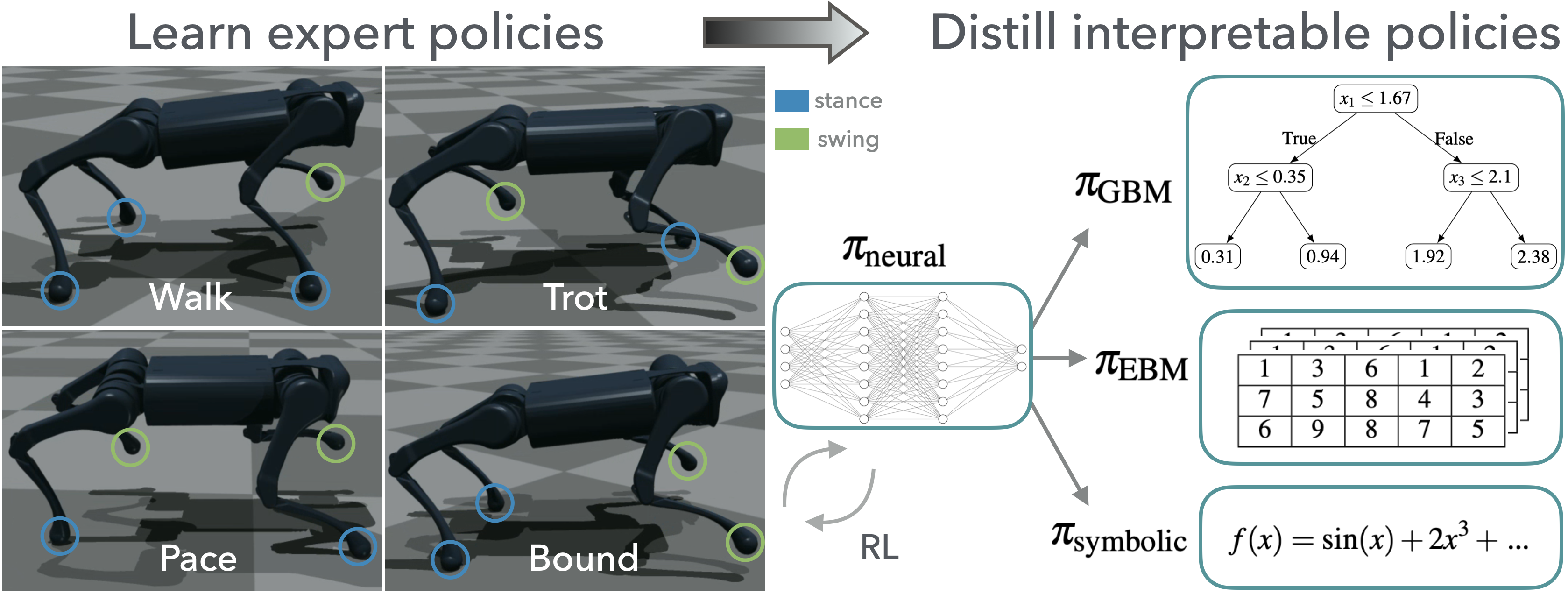}
    \caption{From black-box to glass-box: summary of the proposed framework for distillation of neural network-based RL policies into interpretable policies consisting of GBMs, EBMs, and symbolic policies.}
    \label{fig:distillation}
\end{figure}

RL for robot locomotion has rapidly matured in capabilities in recent years, ranging from early demonstrations of policy gradients for training simple locomotion policies \cite{kohl2004policy}, to the use of animal motion imitation \cite{peng2020learning}, and traversal of challenging terrain \cite{lee2020learning}. Some previous work has focused on developing modular or hierarchical architectures, both in locomotion \cite{yang2020multi, yuan2023hierarchical, yu2023identifying}, and manipulation settings \cite{beyret2019dot, triantafyllidis2023hybrid, hu2023modular}, which are intrinsically not as black-box due to their modular structure, however this was mainly done for improving policy performance or learning efficiency -- without delivering interpretability insights (except for \cite{beyret2019dot} in manipulation). Notably, \cite{yu2023identifying} evaluates observation importance for efficient learning of locomotion policies, but uses neural policies and thus can only use black-box saliency methods for importance analysis. Additionally, recent work has demonstrated the ability to learn exteroceptive policies, from sparse environment perception \cite{acero2022learning, liu2021learning} to more dense or visual perception \cite{miki2022learning, yu2021visual, loquercio2023learning}, further advancing the capabilities of robot locomotion learned via RL -- while maintaining the use of neural networks as policies. 

Nevertheless, there is a growing need to produce interpretable policies and thus enable more widespread adoption of intelligent legged robots. Explainable RL has recently developed in various directions \cite{milani2023explainable}, with policy distillation or extraction becoming increasingly popular: decision trees guided by Q-functions have been distilled from neural policies for simple game environments \cite{bastani2018verifiable}, as well as state machines and list processing programs \cite{bastani2020interpretable}, and decision trees have also been used for evolutionary feature synthesis \cite{zhang2020interpretable} to provide visualizations and rule-based explanations of simple agent-environment interactions \cite{bewley2021tripletree}. Moreover, neural RL expert policies have been distilled into decision trees in various domains where interpretability is crucial, such as power system control \cite{dai2022enhanced}, aircraft separation assurance \cite{guo2022explainable}, and sensor-based robot navigation \cite{roth2021xai}.



To address the need for policy interpretability and inspired by previous work on explainable RL, we develop a novel framework for distilling neural network expert locomotion policies trained via RL into more interpretable glass-box policies, as shown in Figure \ref{fig:distillation}. Our main contributions are:
\begin{itemize} 
\item A novel policy distillation framework incorporating episode-dependent policy alternation to DAgger \cite{ross2011reduction}. 
\item Effective locomotion policies distilled via Gradient Boosting Machines (GBMs) \cite{friedman2001greedy}, Explainable Boosting Machines (EBMs) \cite{lou2012intelligible}, and Symbolic Regression \cite{cranmer2023interpretable}. 
\item Interpretability of the observation-action mapping unveiled in the distilled locomotion policies, and the evaluation of their performance in tasks consisting of walking, trotting, pacing, and bounding gaits, providing both global and local explanations of policy actions.
\end{itemize}
We follow a distillation approach as the interpretable models we use cannot be trained to perform general function approximation parametrically via policy gradients, they are best suited for regression on a supervised dataset. To the best of our knowledge, our work is the first to distill RL locomotion policies into GBMs, EBMs, and symbolic policies.

\section{Background}




\subsection{Reinforcement Learning for Robot Locomotion}
RL is the machine learning paradigm for decision-making or control \cite{sutton_reinforcement_2018}, also known as approximate dynamic programming for solving Markov Decision Processes (MDPs), defined as a tuple $\langle \mathcal{S}, \mathcal{A}, P(s_{t+1} \vert s_{t}, a_{t}), R \rangle$, where $ \mathcal{S}$ is the state space, $ \mathcal{A}$ is action space, $ P(s_{t+1} \vert s_{t}, a_{t})$ is the transition dynamics, and $R$ is the reward function. We denote a policy $\pi:\mathcal{S} \rightarrow \mathcal{A}$ parametrized by $\theta$ as $\pi_{\theta}$. 

The RL objective is to maximize cumulative rewards, and policy gradient algorithms are a popular approach to approximately solve this using differentiable policies $\pi_{\theta}$ such as neural networks, by optimizing an objective of the form:
\begin{equation}
    \nabla_{\theta} \mathbb{E} \left[\sum_{t=0}^{T} r_t \right] \approx \mathbb{E} \left[ \sum_{t=0}^{T} \Psi_t \nabla_{\theta} \log \pi_{\theta}(a_t | s_t) \right]
    \label{eq:policygradient}
\end{equation}
where $\Psi_t$ takes different forms depending on the algorithm, such as discounted returns, temporal-difference residual, or a clipped surrogate objective in the case of the popular algorithm Proximal Policy Optimization (PPO) \cite{PPOSchulman}, which uses the parameter update $\theta_{k+1} = \arg \max_{\theta} \mathbb{E}_{s,a \sim \pi_{\theta_k}} \left[
    L(s,a,\theta_k, \theta)\right]$
where $L(s,a,\theta_k, \theta)$ is a clipped lower bound objective. 


In RL for robot locomotion, the MDP state usually includes joint states, velocities, base orientation, velocity, additional terms like feet height, contact states, target velocity, or distance to target, and exteroceptive information if relevant, with actions typically being joint position targets executed by high-frequency low-level joint PD controllers for compliant behavior \cite{lee2020learning, yang2020multi, yu2023identifying, acero2022learning, miki2022learning, loquercio2023learning}. Reward functions often combine target tracking, joint state or target smoothness, and other shaping terms for desired gaits. Our approach utilises \emph{reward machines} that structure reward functions as state machines and extend the MDP state with the reward machine state, enhancing learning efficiency and locomotion robustness \cite{defazio2024learning}. This also aids policy interpretability through the logical rules of reward machine states. See \cite{defazio2024learning} for an in-depth discussion on locomotion reward machines.

\subsection{Gradient Boosting Machines and Symbolic Regression}

Generalized Additive Models (GAMs) are a flexible class of models that extend linear models by allowing non-linear relationships between each predictor and the response variable, while maintaining additivity \cite{hastie1986generalized}. The model can be expressed as:

\begin{equation}
g(\mathbb{E}[y]) = \beta_0 + f_1(x_1) + f_2(x_2) + \cdots + f_p(x_p)
\label{eq:GAM}
\end{equation}

where $y$ is the response variable, $g(\cdot)$ is a link function (identity for regression, sigmoid for classification), $x_i$ are predictors, $\beta_0$ is the intercept, and $f_i$ are shape functions.

Gradient Boosting Machines (GBMs) are an ensemble learning technique that builds models sequentially, each new model correcting errors made by the previous ones \cite{friedman2001greedy}. A GBM combines weak learner models, typically shallow decision trees, to create a strong predictive model:

\begin{equation}
\hat{y} = \sum_{i=1}^{M} \gamma_i h_i(x) ,
\label{eq:GBM}
\end{equation}
where $\hat{y}$ is the predicted response, $h_i(x)$ are the weak learner models, $\gamma_i$ are the corresponding weights, and $M$ is the number of models.

Explainable Boosting Machines (EBMs) combine the advantages of gradient boosting from GBMs, with the intelligibility of GAMs \cite{lou2012intelligible}. Notably, EBM implementations allow for univariate $f_i$ and optionally bivariate $f_{i,j}$ shape functions when valuable \cite{nori2019interpretml}, expanding GAMs by accounting for pairwise interaction terms as:
\begin{equation}
g(\mathbb{E}[y]) = \beta_0 + \sum f_i(x_i) + \sum f_{i,j} (x_i, x_j)
\label{eq:EBM}
\end{equation}
where $f_i$ and $f_{i,j}$ are essentially learned lookup tables. 

Symbolic regression seeks mathematical models that best describe data, differing from traditional regression by not strictly presupposing the model structure \cite{cranmer2023interpretable}. Utilizing genetic algorithms (GAs), symbolic regression evolves expressions using unary and binary operators to minimize an error metric $\mathcal{L}$ over data $D$ as $\min_f \mathcal{L}(D, f(x))$ where $f(x)$ is usually a GAM with complexity constraints. This symbolic approach enables the discovery of interpretable models, revealing inherent data patterns \cite{cranmer2023interpretable}.

\section{Methodology}
We now present the methodology used in this work. Our framework consists of (i) training of RL experts as neural policies, and (ii) distillation of the neural policies into interpretable policies. We note that we follow this process because the types of interpretable policies we use are not suitable for gradient-based optimization of policy parameters, which is a requirement of policy gradient RL methods. Moreover and noticeably, the neural policies used in previous locomotion work are not particularly deep, usually having 2 to 5 hidden layers \cite{lee2020learning, yang2020multi, acero2022learning}, and hence the limited expressiveness of these networks suggests that the observation-action mapping learned via RL can be distilled into simpler forms, such as decision trees or additive models, which motivates our work. 

\begin{table*}[ht]
\centering
\caption{Top 3 Observations by Importance for Each Joint Type Action From Distilled Gradient Boosting Machine Policies Based on Feature and Permutation Importance for 4 Different Gaits}
\label{tab:top_features}
\scalebox{0.90}{\begin{tabularx}{2\columnwidth}{@{}llXXX@{}}
\toprule
Gait & Joint Type & 1st Feature & 2nd Feature & 3rd Feature \\ \midrule
\multicolumn{5}{c}{Feature Importance} \\ \midrule
Walk & Hip & Prev Action Hip & RM State & RM Iters, DoF Pos or Prev Action Calf, Height Foot  \\
 & Thigh & Prev Action Thigh & RM State, RM Iters, DoF Pos Thigh  & Prev Action Hip, Command X \\
 & Calf & Prev Action Hip or Calf & RM Iters, Prev Action Hip or Calf & Prev Action Hip or Thigh \\
Trot & Hip & Prev Action Hip & RM State, RM Iters, Prev Action Hip or Thigh & RM Iters, Prev Action Calf \\
 & Thigh & Prev Action Hip or Thigh & RM Iters, Command X, Prev Action Thigh or Calf & RM Iters, RM State, Foot Height \\
 & Calf & Prev Action Hip or Calf & Prev Action Calf or Hip, RM State & Prev Action Hip or Thigh \\
Pace & Hip & Prev Action Hip, Dof Vel Hip, RM Iters, Foot Height & DoF Pos or Vel Thigh, RM State, RM Iters & Prev Action Hip, Foot Height, DoF Vel Calf \\
 & Thigh & Prev Action Hip, RM Iters, Foot Height & Foot Height, Prev Action Hip & DoF Vel Hip or Thigh, Command X \\
 & Calf & Prev Action Hip or Thigh & Prev Action Thigh or Calf, RM Iters, Foot Height & RM Iters, DoF Pos or Vel Thigh \\
Bound & Hip & Prev Action Hip & Prev Action Hip or Thigh & Prev Action Calf, Base Lin Vel Z \\
 & Thigh & Prev Action Thigh & Prev Action Thigh or Calf, RM Iters & Foot Height, Command X, Prev Action Hip or Calf \\
 & Calf & Prev Action Calf or Hip & Prev Action Calf or Hip & Prev Action Hip or Thigh \\ \midrule
\multicolumn{5}{c}{Permutation Importance} \\ \midrule
Walk & Hip & Prev Action Hip & RM State, Foot Height & RM Iters, RM State, Prev Action Calf \\
 & Thigh & Thigh Prev Action & RM State, RM Iters, Command X  & Prev Action Hip, Command X \\
 & Calf & Prev Action Hip or Calf & Prev Action Calf or Hip, RM State & Prev Action Hip or Thigh \\
Trot & Hip & Prev Action Hip & RM State, RM Iters, Prev Action Hip or Calf & RM Iters, Prev Action Thigh or Calf \\
 & Thigh & Prev Action Hip or Thigh & RM Iters, Command X, Prev Action Thigh or Calf & RM Iters, RM State, Foot Height \\
 & Calf & Prev Action Hip or Calf & Prev Action Calf or Hip, RM State & Prev Action Hip or Thigh \\ 
Pace & Hip & Foot Height, Prev Action Hip, RM Iters & Prev Action Hip, DoF Pos Hip, RM Iters, DoF Vel Calf & Prev Action Hip or Thigh, DoF Vel Thigh, RM State \\
 & Thigh & Prev Action Hip, RM Iters, Foot Height & Foot Height, Prev Action Hip & DoF Vel Hip or Thigh, Command X \\
 & Calf & Prev Action Hip or Thigh & Prev Action Thigh or Calf, RM Iters, Foot Height & RM Iters, DoF Pos or Vel Thigh \\
Bound & Hip & Prev Action Hip & Prev Action Hip or Thigh & Prev Action Calf, RM State \\
 & Thigh & Prev Action Thigh & Command X, Prev Action Hip & RM State, RM Iters, Prev Action Thigh or Calf \\
 & Calf & Prev Action Calf or Hip & Prev Action Calf or Hip, Base Lin Vel Z & Prev Action Hip or Calf \\ 
\bottomrule
\end{tabularx}}
\end{table*}

\subsection{Training Reinforcement Learning Expert Policies}



In general, RL algorithms require $\pi_{\theta}$ to be differentiable, and therefore we cannot easily train GBMs, EBMs, or symbolic policies directly via RL as these are not directly amenable for gradient-based optimization. Thus, we use neural networks for our experts. 
We train expert policies for the following tasks or gaits: walk, trot, pace, and bound. To obtain our neural expert policies, we use the PPO algorithm in IsaacGym simulations. As previously mentioned, we build on top of \cite{defazio2024learning}, but we note that our approach is agnostic to the specific details regarding the RL methodology used to train the neural experts. The logical propositions used for defining the reward machines for each gait can be found in \cite{defazio2024learning}. All gaits use the same base observations and only differ in their reward machine states. Full observation lists are in Figure \ref{fig:impsall}. A learned state estimator is used for base velocity and feet contact forces \cite{defazio2024learning}, but other alternatives could be used. 

We use the Unitree A1 quadruped robot for our experiments. Each expert policy is trained with randomized forward velocity commands in the range $[-1,1]$ m/s and yaw rate commands in $[-1,1]$ rad/s. The control frequency is 50Hz, with joint PD controllers set at $P=20$ and $D=0.5$ as \cite{defazio2024learning}. We train the expert policy for each gait for 1.5k PPO updates \cite{PPOSchulman} using 1024 parallel environments, which equates to approximately \textbf{205 hours} of simulated time used to train each expert -- substantially less than previous work \cite{acero2022learning}, highlighting the sample efficiency of using reward machines. 

\subsection{Distilling Interpretable Locomotion Policies}
In essence, our distillation process is an imitation learning problem, where the expert policies have been trained via RL. Therefore, it is subject to the distribution shift found in vanilla behavioural cloning. To address this, we use the Dataset Aggregation method (DAgger) \cite{ross2011reduction}. However, instead of directly combining pure expert and imitation policy rollouts in the supervised dataset, we modify DAgger to use episode-dependent alternation of actions given by the expert and distilled policies, as shown in Algorithm \ref{alg:ourdagger}. We experimentally found that without this modification policy performance was poor, yielding unstable gaits (this might be addressable by increasing $max\_episodes$ substantially, but it could make distillation prohibitively expensive). 

The \textit{alternation ratio} $1/n$ determines how often the expert actions are used during rollouts, with $n$ increasing during the distillation process as a curriculum. This modification is well motivated for robot control settings with feedback policies, as the action alternation leads to a more graceful trajectory distribution shift in the data used to train the distilled policy. 


We used $t=1000$, corresponding to only \textbf{10 minutes} as the total simulated time in the distillation dataset $D$ for each gait. Specifically, policies are trained with $max\_episodes=30$ alternating linear velocity commands in $[0, 0.25, 0.5, 0.75]$ and only updating $n$ after cycling through all the velocity commands (i.e. $n_f = 4$), thus the lowest alternation ratio found in the datasets used for distillation is $1/8$.

Using Algorithm \ref{alg:ourdagger}, in step 19 we distill three types of interpretable locomotion policies for each gait: GBMs, EBMs, and symbolic expressions, leaving 20\% of $D$ as test set. We leverage efficient implementations from \cite{pedregosa2011scikit} for GBMs, from \cite{nori2019interpretml} for EBMs, and from \cite{cranmer2023interpretable} for Symbolic Regression, with default hyperparameters for each as they are optimized for robust performance. For Symbolic Regression, we use the unary operators $[ \sin(\cdot), \tanh(\cdot), \cdot^2, \cdot^3]$, and binary operators $[\cdot + \cdot, \cdot - \cdot, \cdot \times \cdot]$ with maximum operator complexity 4 and overall complexity 90, for 20 iterations per distillation.

\begin{algorithm}[t]
\caption{DAgger with Curriculum of Episode-Dependent Alternation of Expert and Distilled Policy Actions}
\begin{algorithmic}[1]
\State Initialize dataset $D \gets \emptyset$
\State Initialize distilled policy $\pi_{\text{distilled}}$ randomly
\State Initialize pre-trained expert policy $\pi_{\text{expert}}$
\State Set frequency parameter $n_f$
\State Set maximum episodes $max\_episodes$
\For{$episode = 1$ \textbf{to} $max\_episodes$}
    \State Set $n \gets \max(1, \lceil episode / n_f \rceil)$
    \State Initialize episode trajectory $\tau \gets \emptyset$
    \For{each step $t$ of the episode}
        \If{$t \mod n = 0$}
            \State Execute action $a_t \gets \pi_{\text{expert}}(s_t)$ \Comment{Use expert policy every $1/n$ steps}
        \Else
            \State Execute action $a_t \gets \pi_{\text{distilled}}(s_t)$ \Comment{Use distilled policy otherwise}
        \EndIf
        \State Observe new state $s_{t+1}$ and reward $r_t$
        \State Append $(s_t, a_t, s_{t+1}, r_t)$ to $\tau$
    \EndFor
    \State Aggregate dataset $D \gets D \cup \{(s_t, \pi_{\text{expert}}(s_t)) \,|\, (s_t, \cdot, \cdot, \cdot) \in \tau\}$
    \State Update $\pi_{\text{distilled}}$ by supervised learning on $D$
\EndFor
\end{algorithmic}
\label{alg:ourdagger}
\end{algorithm}
\vspace{-0mm}

\section{RESULTS}
We present the results of the GBM, EBM, and symbolic policies across various gaits: walking, trotting, pacing, and bounding. A comprehensive data analysis is conducted to thoroughly delineate both the performance and interpretability of these policies.

\subsection{Performance of Distilled Policies}
We evaluate the performance of all distilled policies after termination of Algorithm \ref{alg:ourdagger}. We do this from the perspective of regression performance and task performance during policy rollouts in our simulated environment.

\begin{table}[ht]
\centering
\caption{Comparison of \(R^2\) Scores on Test Sets Across Different Gaits}
\label{tab:r2_scores}
\scalebox{1.0}{\begin{tabularx}{\columnwidth}{@{}XXXXX@{}}
\toprule
Model Type & Walk & Trot & Pace & Bound \\ \midrule
GBM & 0.9705 & 0.9863 & 0.9752 & 0.9537 \\
EBM & 0.9787 & 0.9906 & 0.9819 & 0.9637 \\
Symbolic & 0.6811 & 0.7334 & 0.7331 & 0.6564 \\ \bottomrule
\end{tabularx}}
\end{table}

\begin{figure}[ht]
    \centering
    \includegraphics[width=\columnwidth]{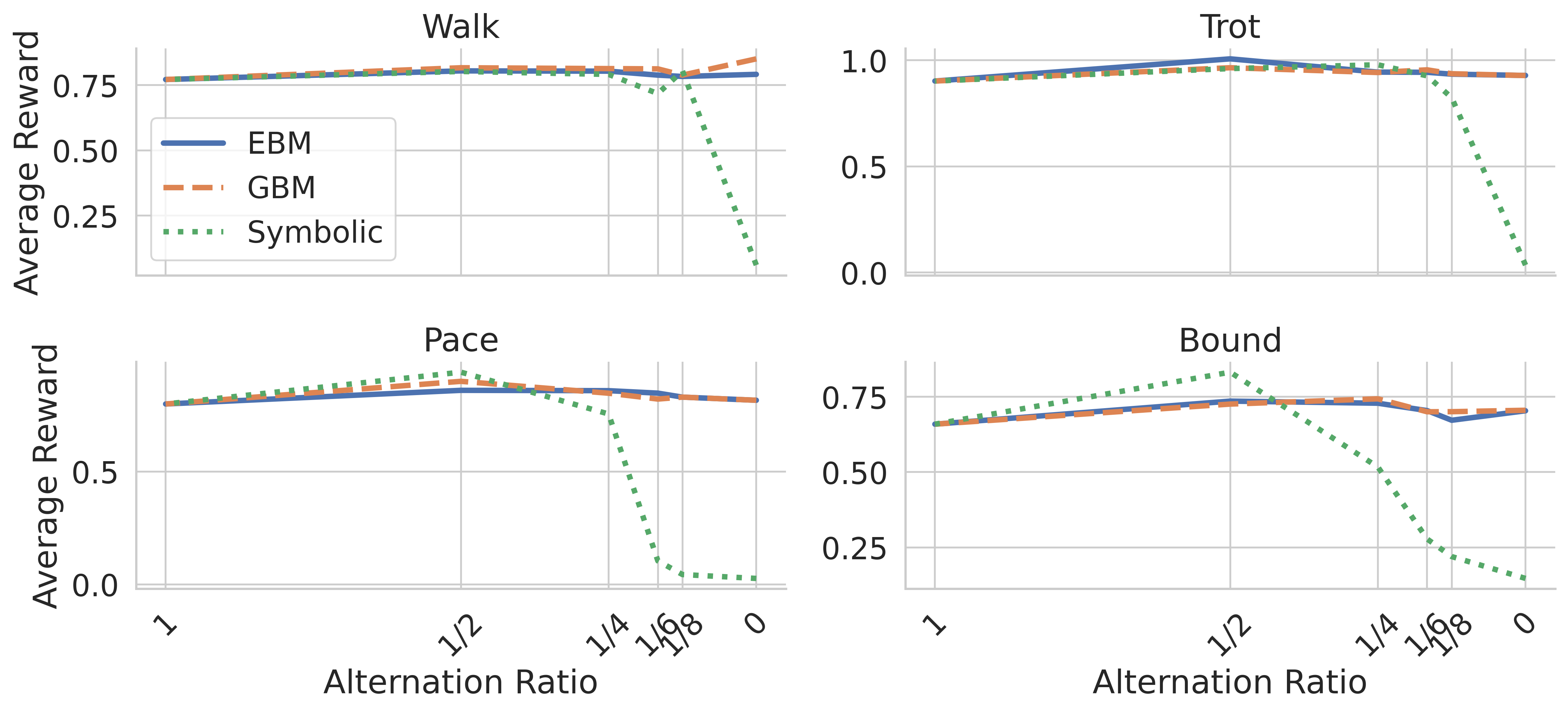}
    \caption{Average episodic performance of distilled policies for all gaits tested using various alternation ratios. Note that alternation ratio of 1 means only the neural RL expert is used, 0 means only the distilled policy is used.}
    \label{fig:average-rewards}
\end{figure}

The regression performance of each method at imitating the corresponding expert policies for each gait quantified by the $R^2$ score is shown in Table \ref{tab:r2_scores}. We note how EBMs and GBMs perform similarly for all gaits, with EBMs performing best, and symbolic policies performing worst. Performance of the symbolic policy might be improved if the genetic algorithm were to be run for more iterations, however these are significantly time-consuming to run and we present results of the best performing unary and binary operators we found after testing various combinations. 

We evaluate each distilled policy upon termination of Algorithm \ref{alg:ourdagger} in simulation with 26 parallel environments using various alternation ratios and provide average episodic rewards in Figure \ref{fig:average-rewards}. These results provide several relevant insights. First, GBM and EBM policies generally maintain performance regardless of the alternation ratio used, whereas symbolic policies yield degraded performance as the neural RL expert is used less often, which is aligned with scores in Table \ref{tab:r2_scores}. Second, it shall be noted that for all gaits there is at least one configuration that outperforms the RL expert (i.e. alternation ratio of 1). Notably, when used strictly by themselves (i.e. alternation ratio 0), the GBM walk policy outperforms the neural RL expert by over 10\%, the EBM and GBM trot policies by 3\%, the EBM and GBM pace policies by 2\%, and the EBM and GBM bound policies by nearly 7\%. This is usually due to better linear and angular velocity reward performance. The performance of the symbolic policies generally matches and sometimes outperforms alternatives when alternated with RL experts (by 12\% for pace and 10\% for bound with alternation ratio $1/2$), but decays rapidly for alternation ratios below $1/6$ for walk and trot, and below $1/4$ for pace and bound, yielding unusable policies in isolation. It shall be noted standalone evaluations of distilled policies (i.e. alternation ratio of 0) constitute a setting that was never seen in the distillation training data. 

\begin{figure}[h]
    \centering
    \includegraphics[width=1.\columnwidth]{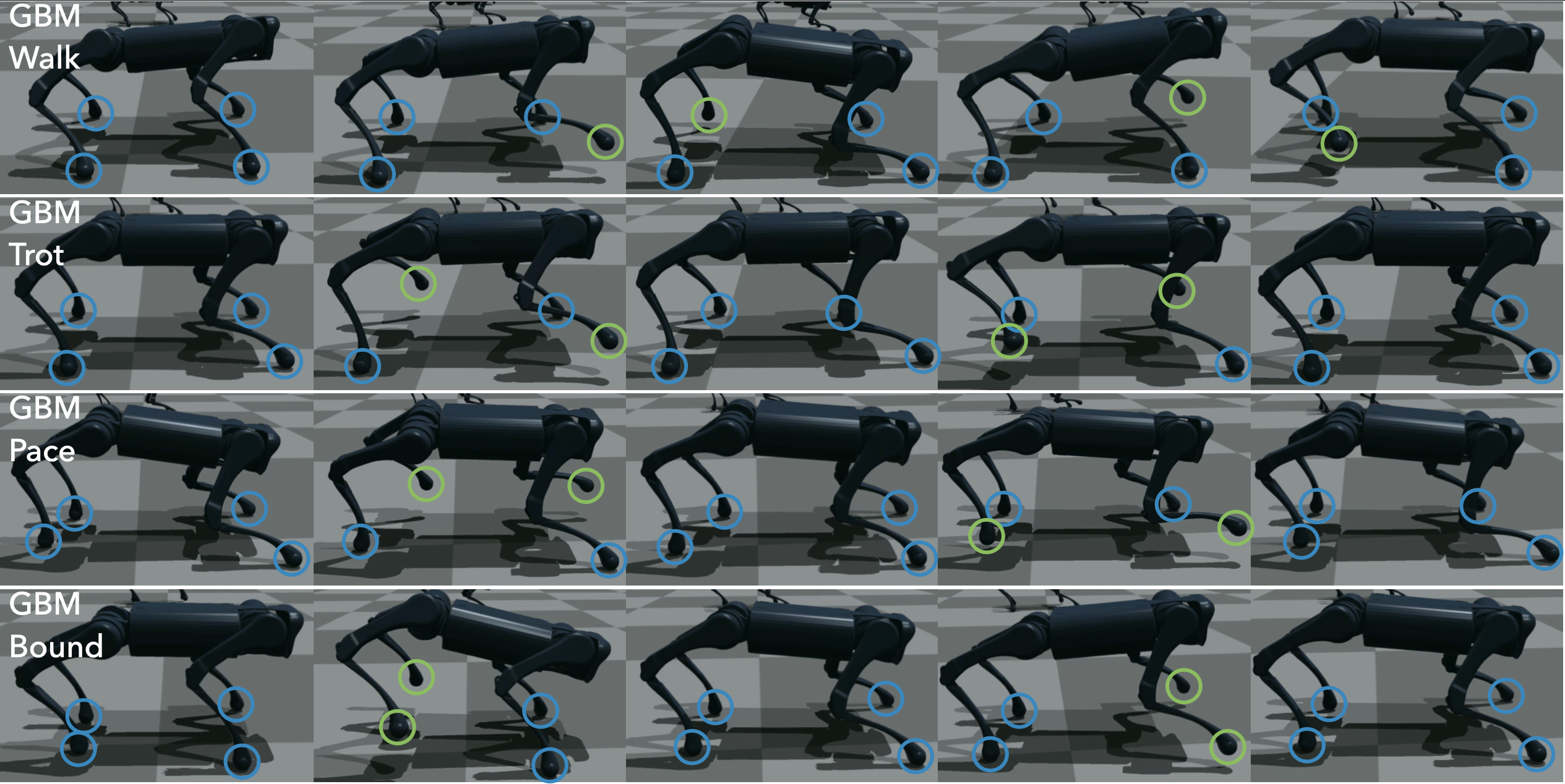}
    \caption{Gait sequences for walk, trot, pace, and bound GBM policies.}
    \label{fig:gbm-all}
\end{figure}

\begin{figure}[h]
    \centering
    \includegraphics[width=1.\columnwidth]{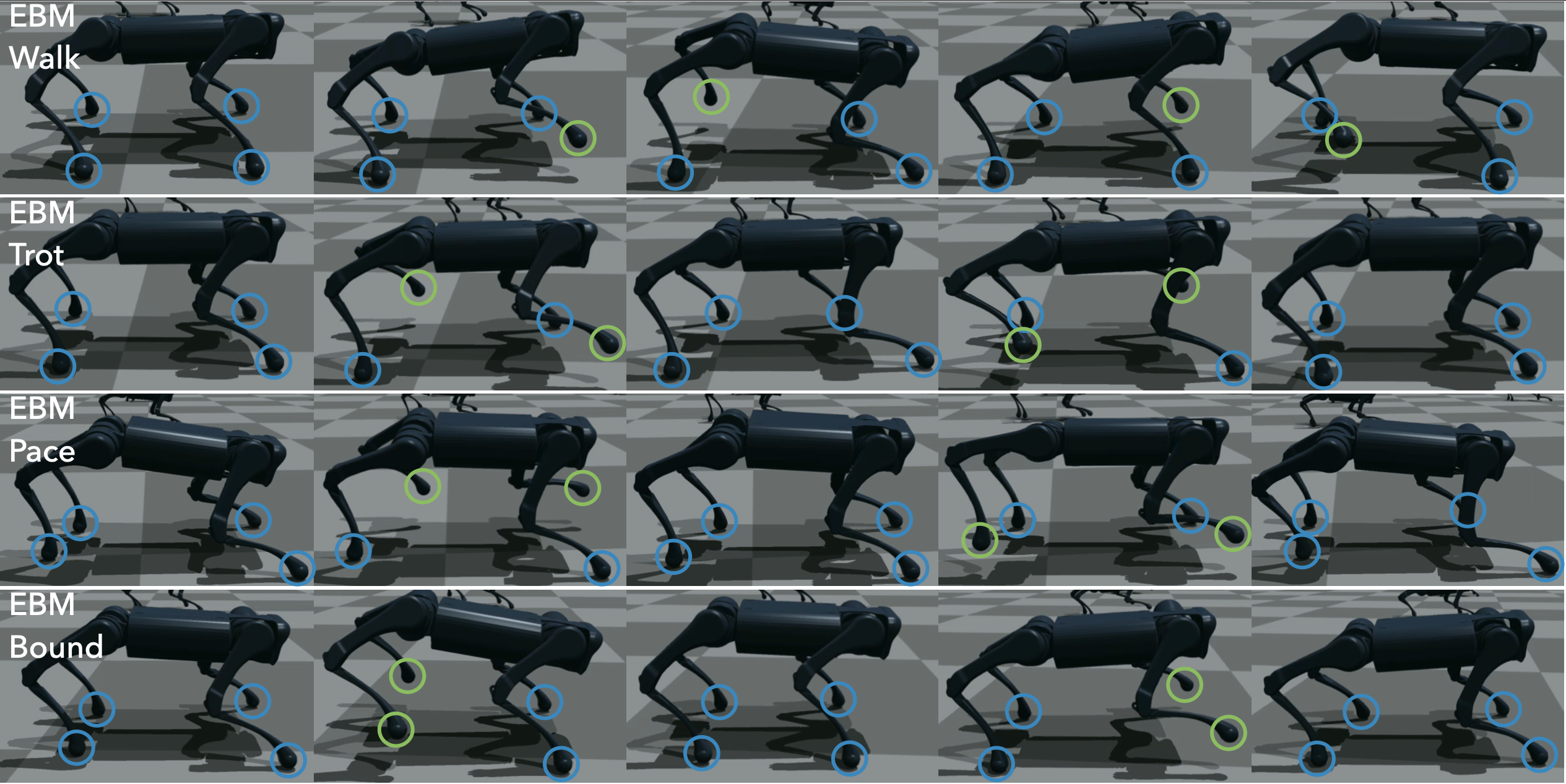}
    \caption{Gait sequences for walk, trot, pace, and bound EBM policies.}
    \label{fig:ebm-all}
\end{figure}

\begin{figure}[h]
    \centering
    \includegraphics[width=1.\columnwidth]{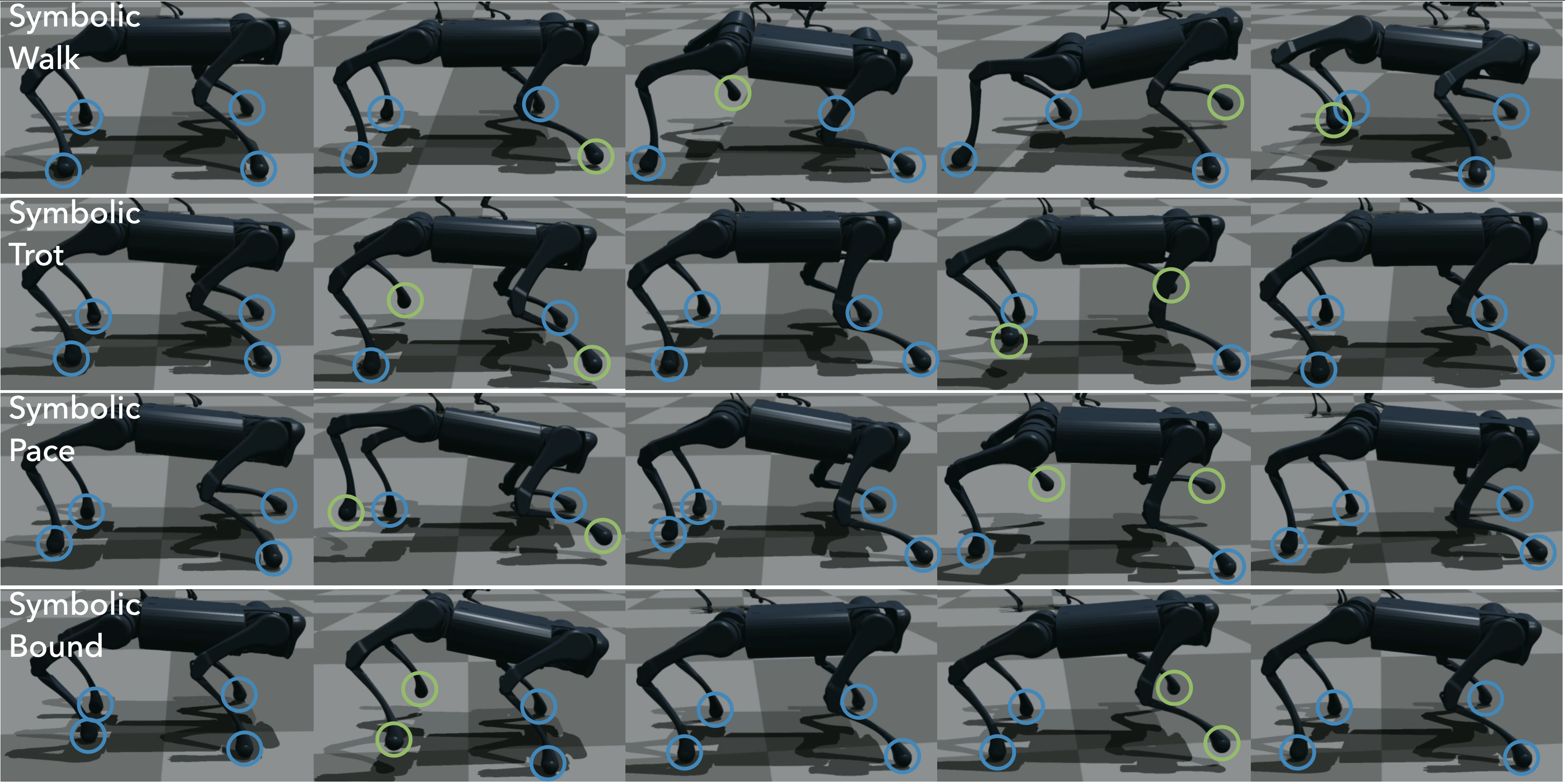}
    \caption{Gait sequences for walk, trot, pace, and bound symbolic policies.}
    \label{fig:sym-all}
\end{figure}

Additionally, we provide a visual depiction of the gait sequences when testing the distilled policies running in isolation (i.e. alternation ratio 0), with GBM policies shown in Figure \ref{fig:gbm-all}, EBM policies in Figure \ref{fig:ebm-all}, and symbolic policies in Figure \ref{fig:sym-all}. It shall be noted how GBM and EBM policies yield visually similar gaits, whereas the symbolic policy yields visibly worse gaits, which is aligned with the results in Figure \ref{fig:average-rewards} and Table \ref{tab:r2_scores}. Tested in isolation, only GBM and EBM policies yielded stable gaits that could run for the full duration of the test episodes, whereas symbolic gaits were not able to sustain more than a couple of gait cycles. 

\vspace{-2mm}

\vspace{-2mm}

\vspace{-2mm}

\begin{figure}[h]
    \centering
    \includegraphics[width=\columnwidth]{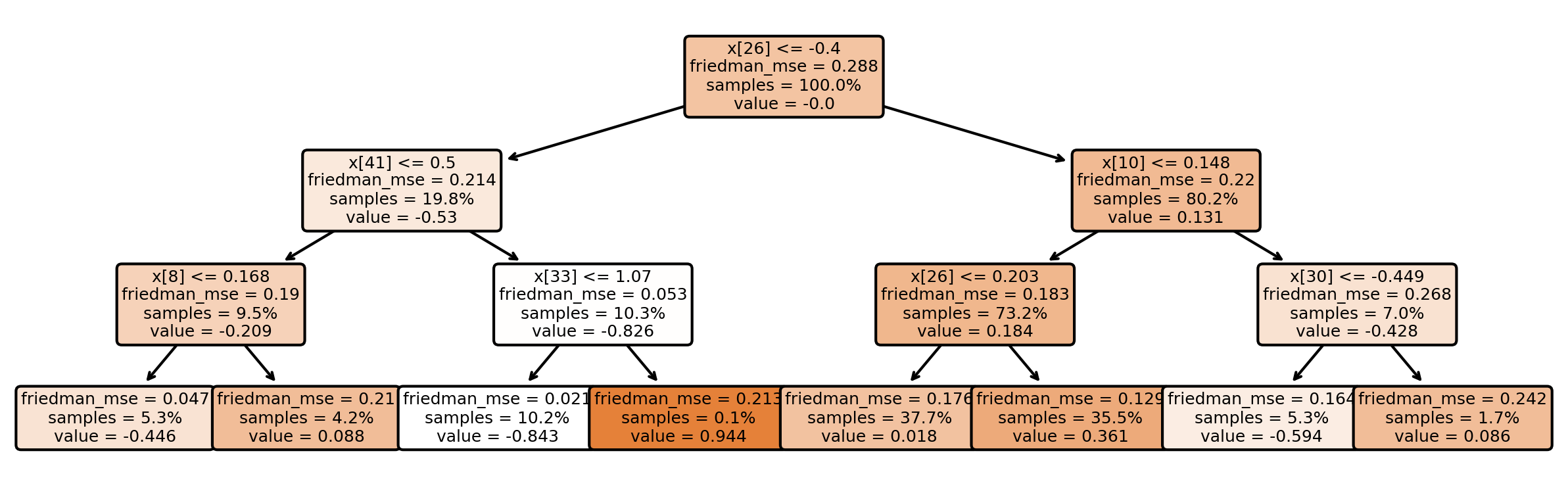}
    \caption{Example of one of the decision trees used as weak learners in the distilled GBM walk policy for Front Left Hip.}
    \label{fig:decision-tree-walk}
\end{figure}

\begin{figure}[h]
    \centering
    \includegraphics[width=0.99\columnwidth]{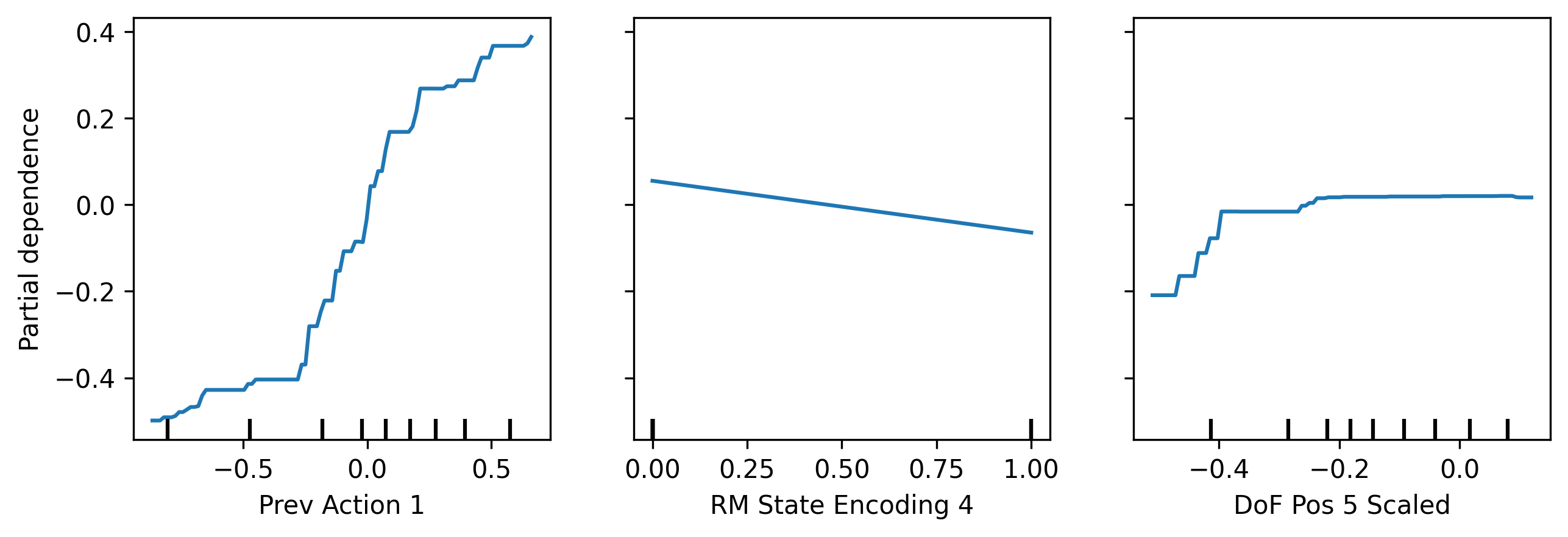}
    \caption{Partial dependence to top 3 observations by feature importance in distilled GBM walk policy for Front Left Hip.}
    \label{fig:partial-dep-walk}
\end{figure}

\begin{figure}[h]
    \centering
    \includegraphics[width=1.00\columnwidth]{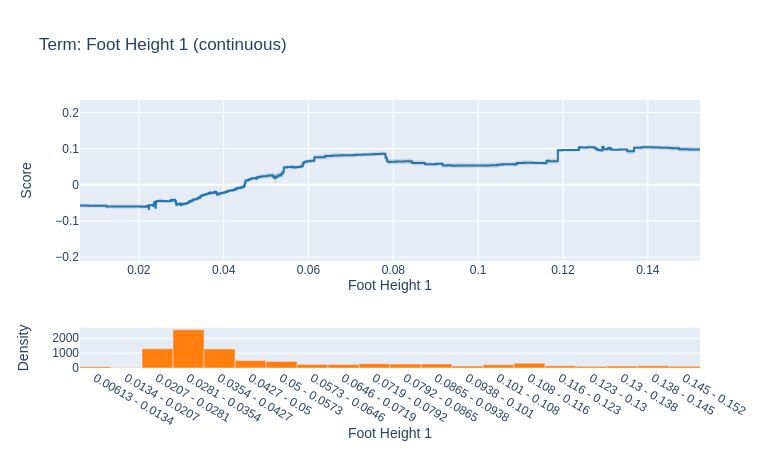}
    \includegraphics[width=1.00\columnwidth]{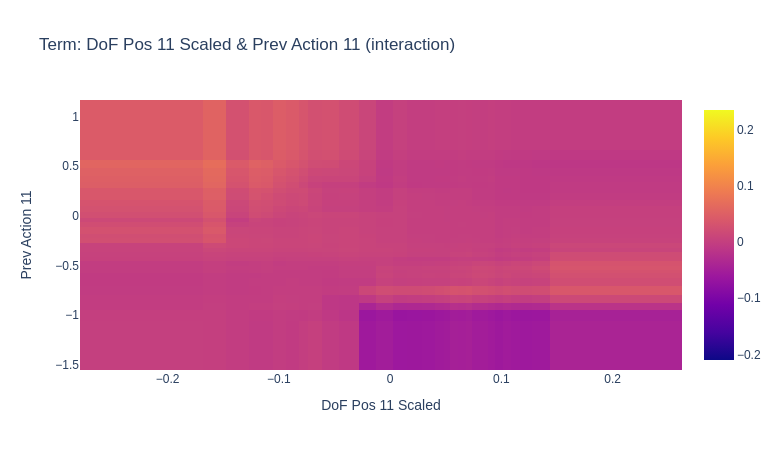}
    \caption{Example global explanations (top: single observation, bottom: interacting observation pair) for EBM trot policy actions for Front Left Hip joint.}
    \label{fig:ebm-trot-feats}
\vspace{-2mm}
\end{figure}

\begin{figure}[h]
    \centering
    \includegraphics[width=1.05\columnwidth]{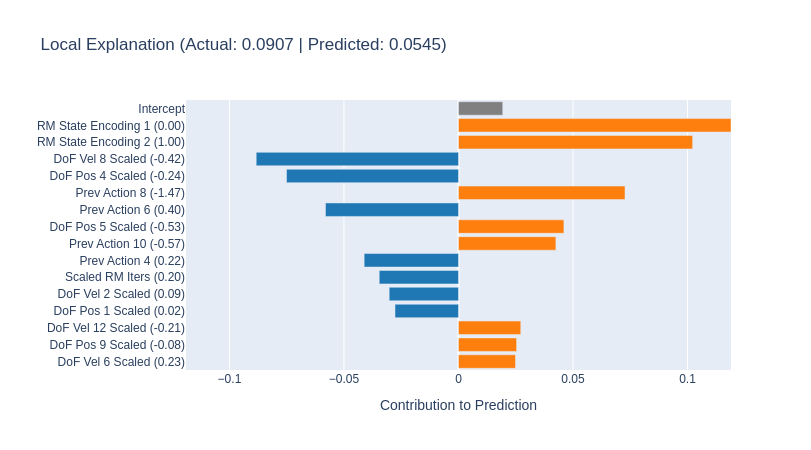}
    \caption{Example local explanation for EBM pace policy action for Front Left Hip joint from an evaluation rollout.}
    \label{fig:ebm-pace-local}
\vspace{-2mm}
\end{figure}

\subsection{Interpretability of Distilled Policies}

\begin{figure*}[]
    \centering
    \includegraphics[width=0.67\columnwidth]{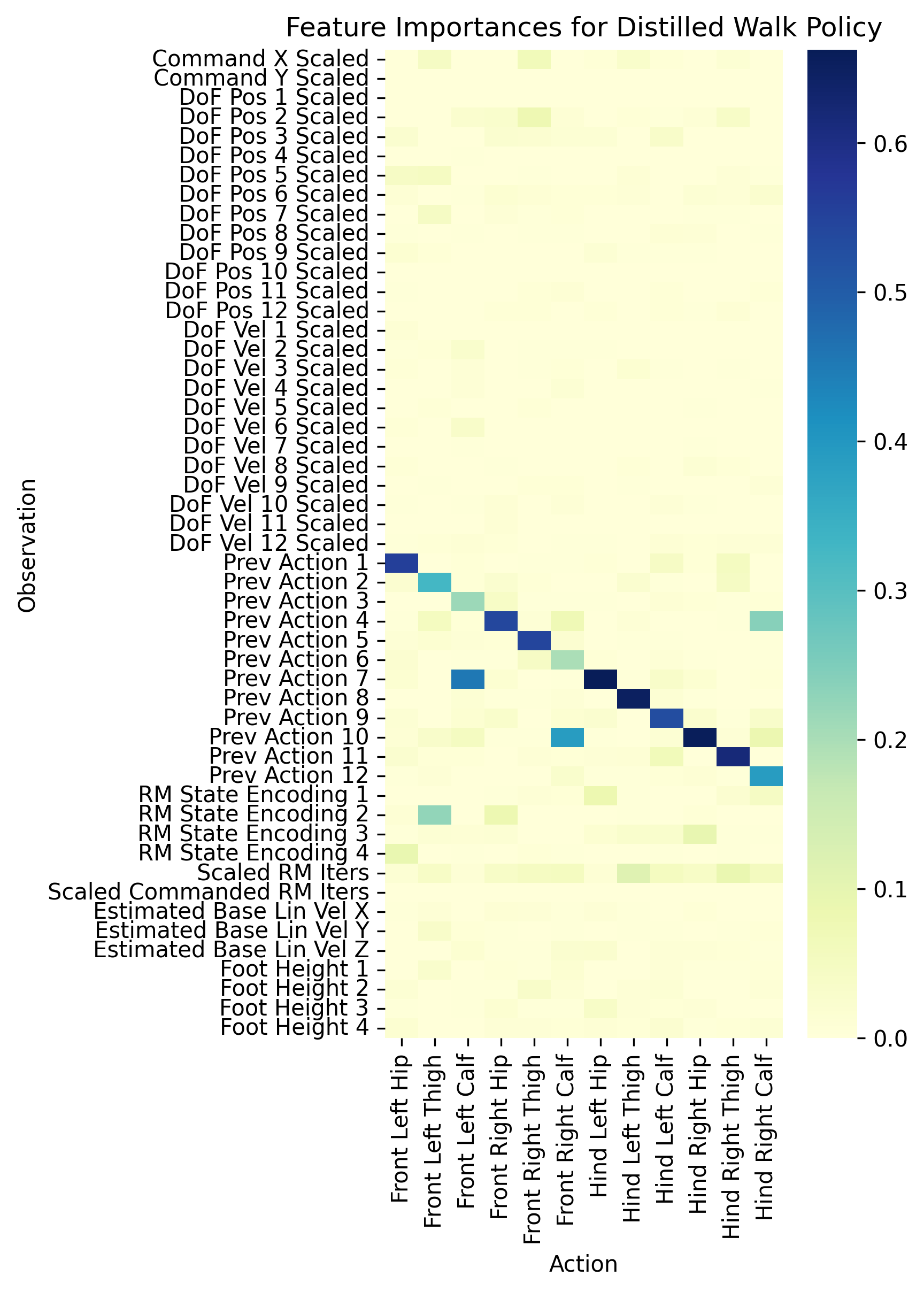}
    \includegraphics[width=0.67\columnwidth]{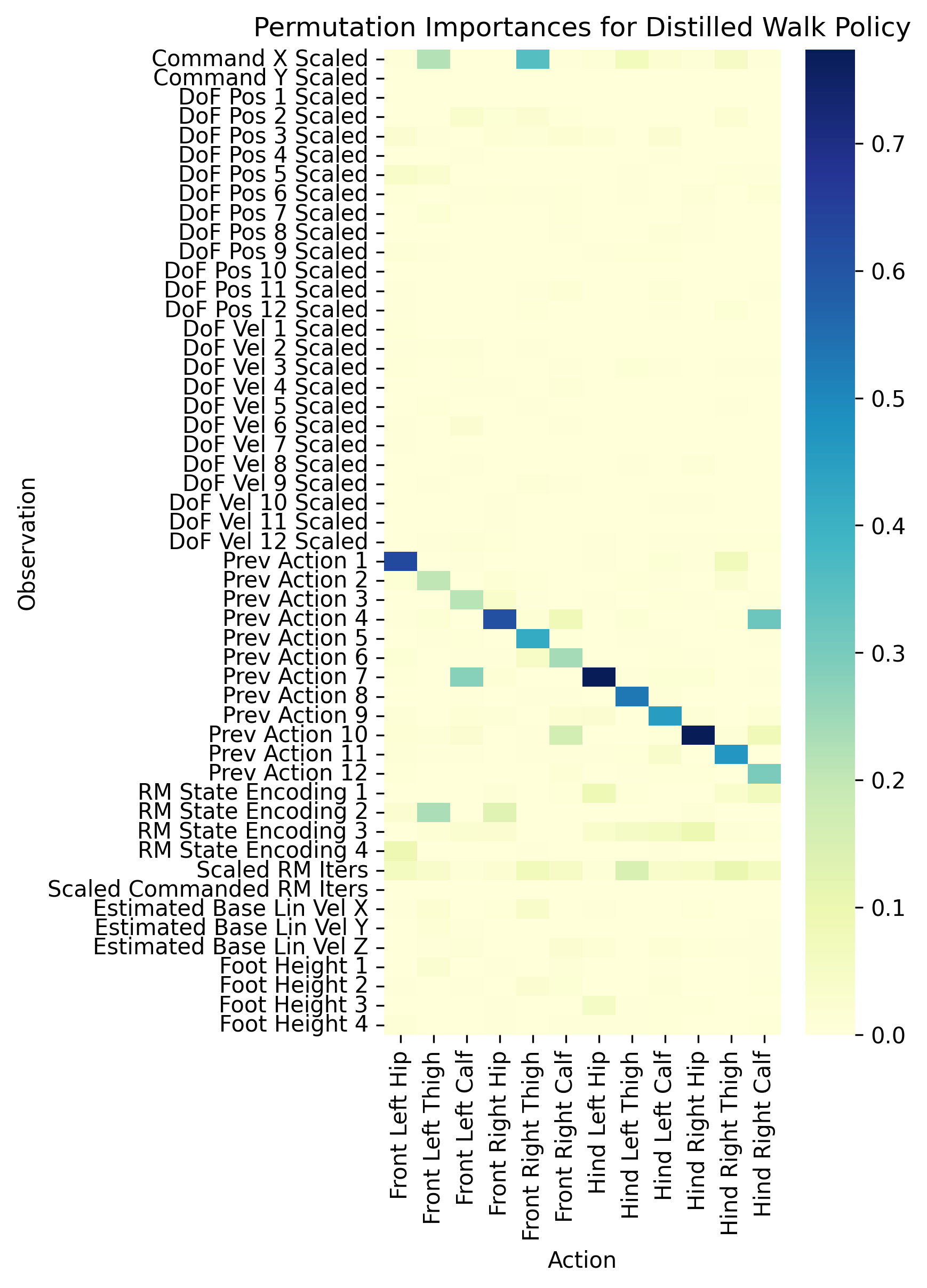}
    \includegraphics[width=0.67\columnwidth]{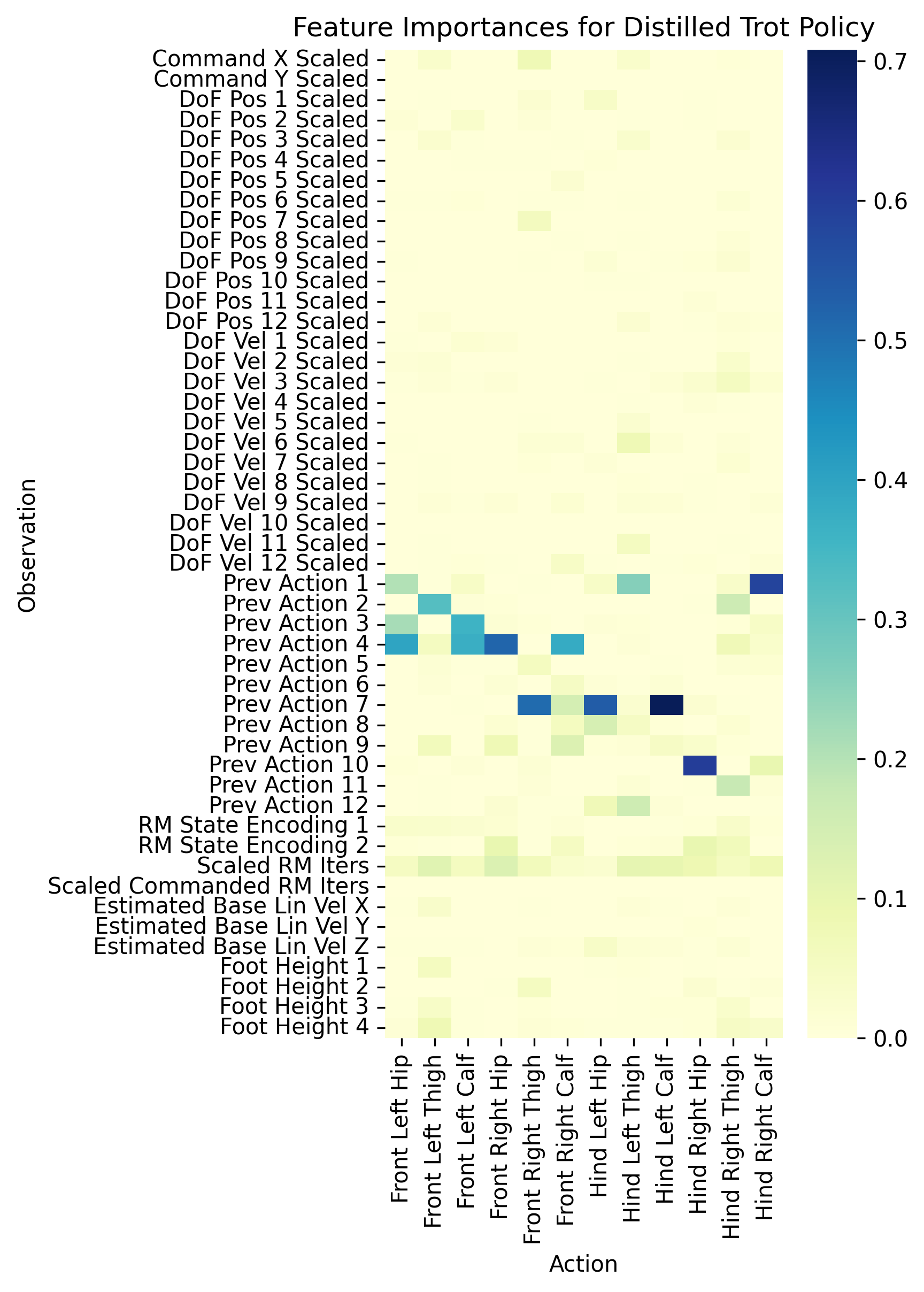}
    \includegraphics[width=0.67\columnwidth]{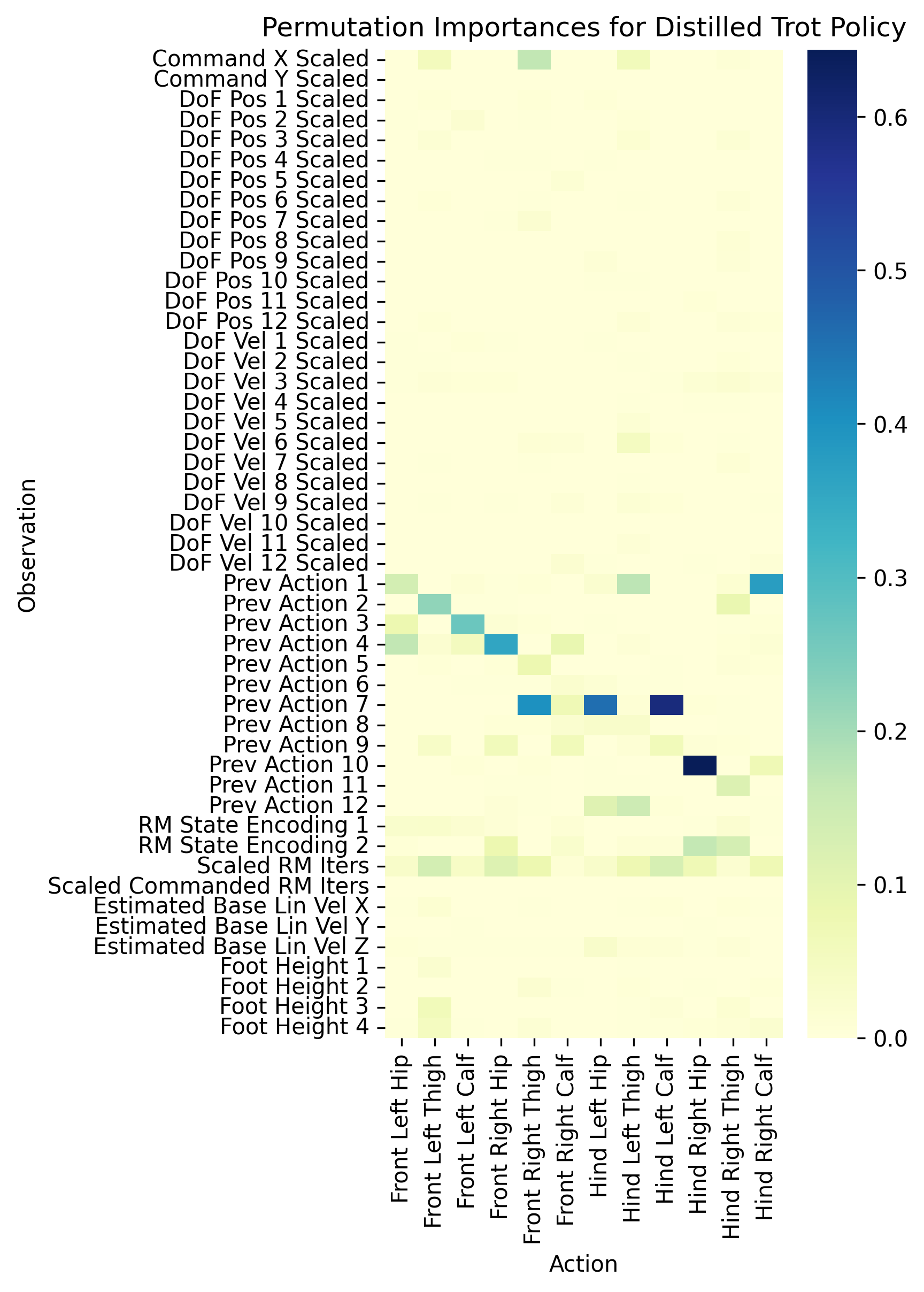}
    \includegraphics[width=0.67\columnwidth]{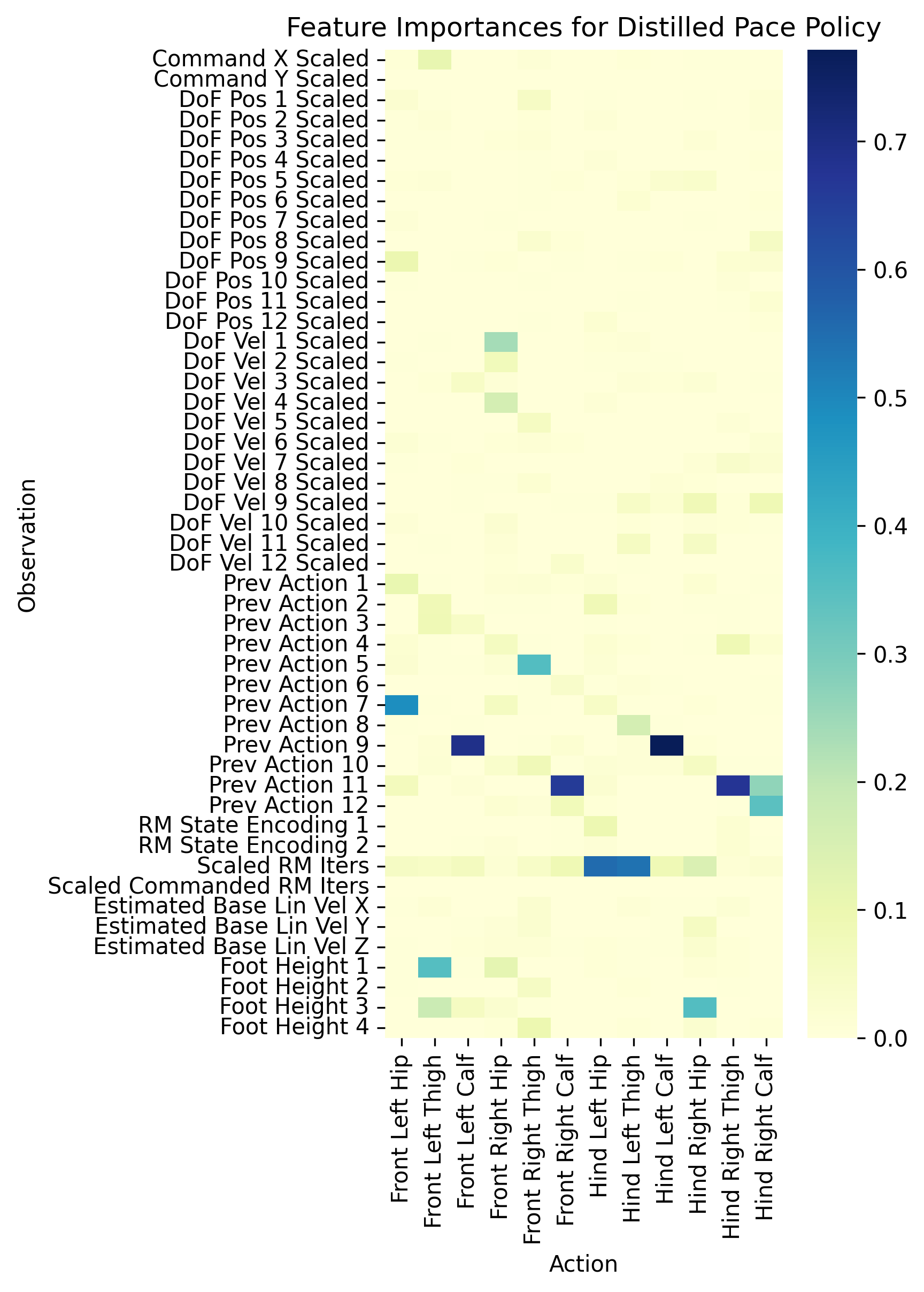}
    \includegraphics[width=0.67\columnwidth]{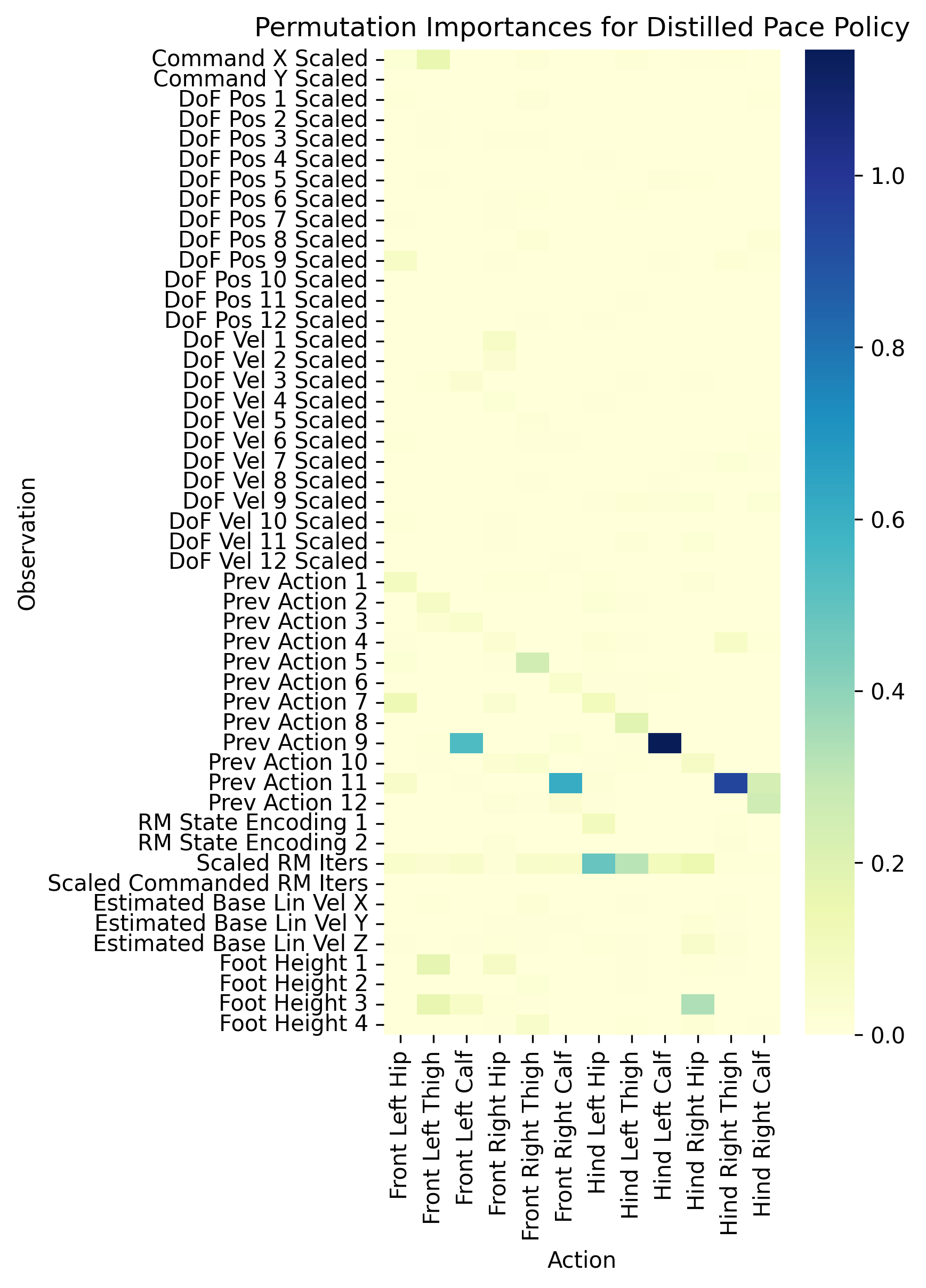}
    \includegraphics[width=0.67\columnwidth]{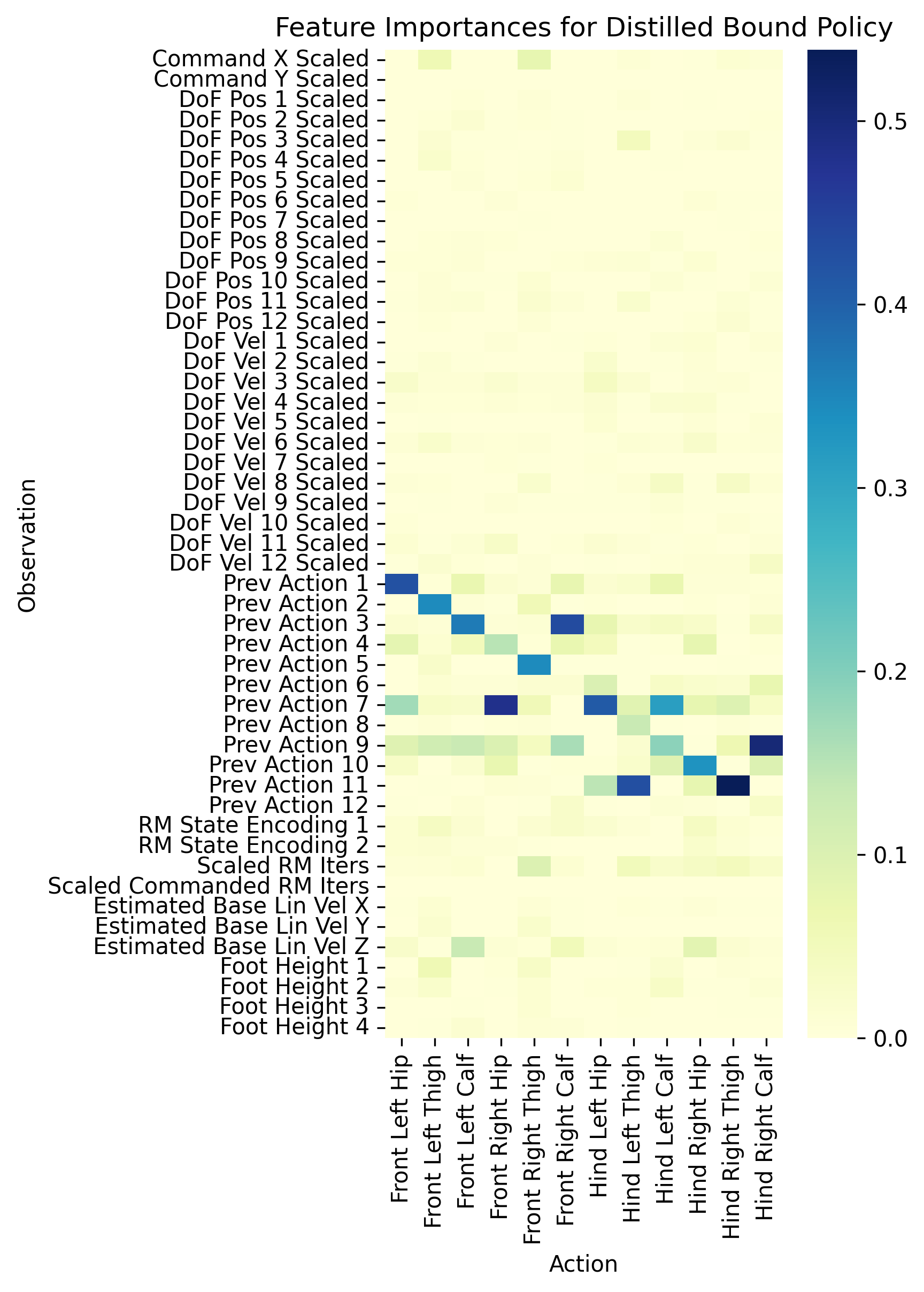}
    \includegraphics[width=0.67\columnwidth]{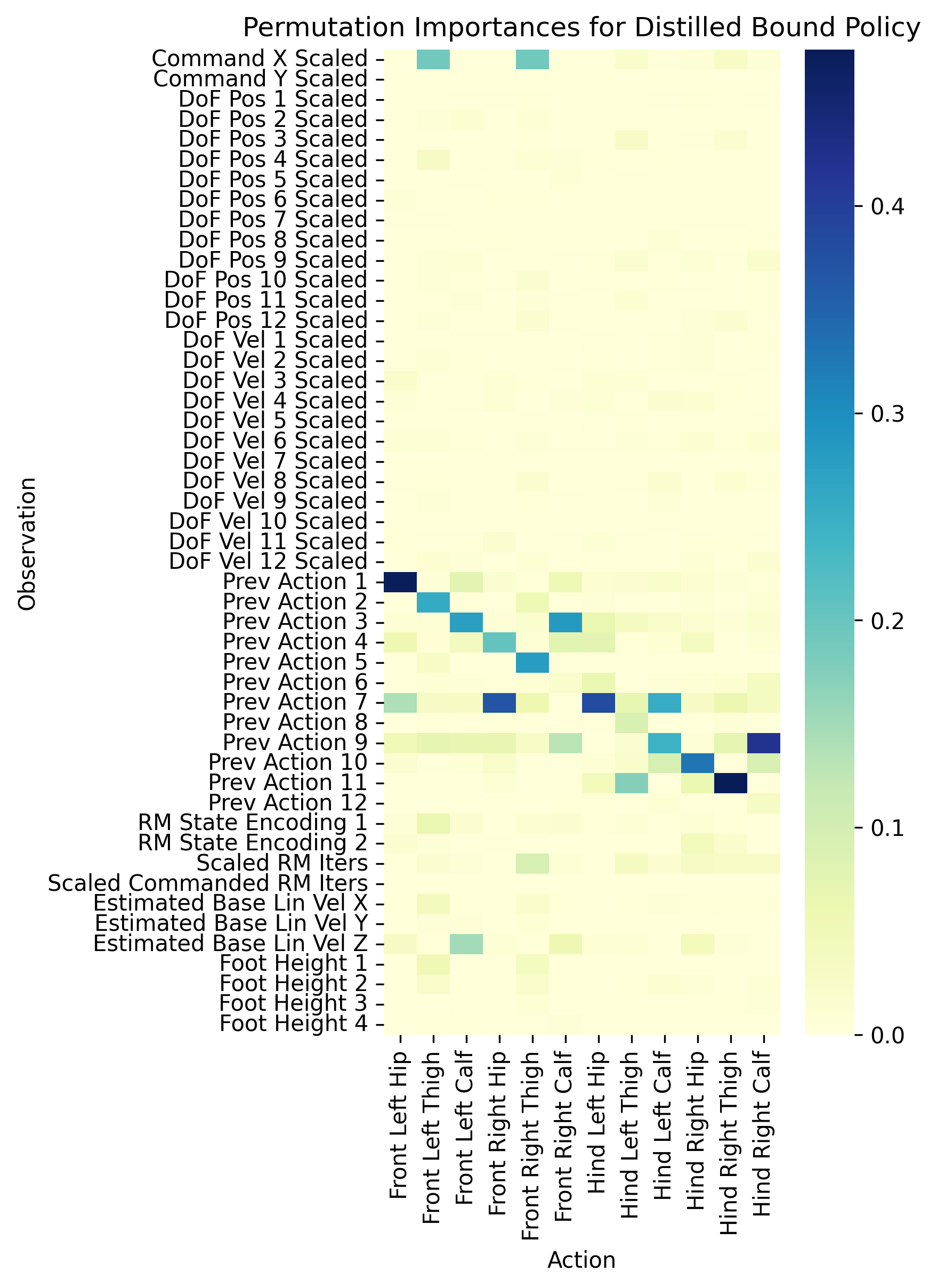}
    \caption{Observation importances for GBM policies computed using two distinct methods: feature importance and permutation importance \cite{pedregosa2011scikit}. }
    \label{fig:impsall}
\end{figure*}

With regards to GBMs, we use two different methods for policy interpretability: feature importance and permutation importance \cite{pedregosa2011scikit}, which quantify importance based on decision tree branches and the effect of permutations respectively. We provide the importance maps for all gaits in Figure \ref{fig:impsall}, and we also provide a summary of those results based on joint type (hip, thigh, calf) for the top 3 observations for each method and gait in Table \ref{tab:top_features}. We note how generally the differences between importance methods is found on the third or second most relevant feature, mostly agreeing on the top feature for all joint types. These results provide a decomposition of the observations relevant for producing the behaviour corresponding to each gait for each joint level in quadruped locomotion. GBMs allow for some global explanations via inspection of decision trees or partial dependence plots as shown in Figures \ref{fig:decision-tree-walk} and \ref{fig:partial-dep-walk} from which counterfactual information could be obtained for certification purposes, e.g. what value should an observation have taken for an action output to be beyond a certain level.

\begin{figure}[]
    \centering
    \includegraphics[width=1.09\columnwidth]{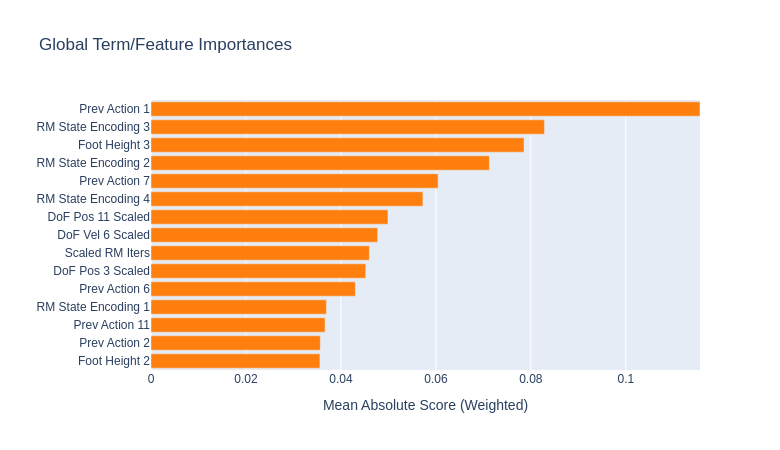}
    \includegraphics[width=1.09\columnwidth]{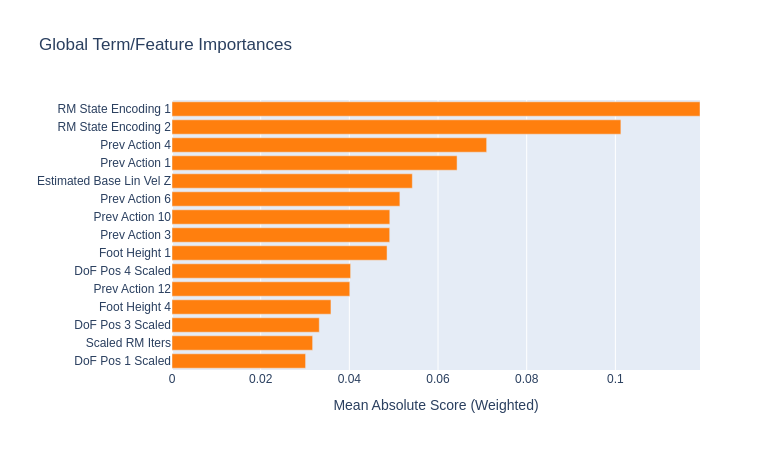}
    \includegraphics[width=1.09\columnwidth]{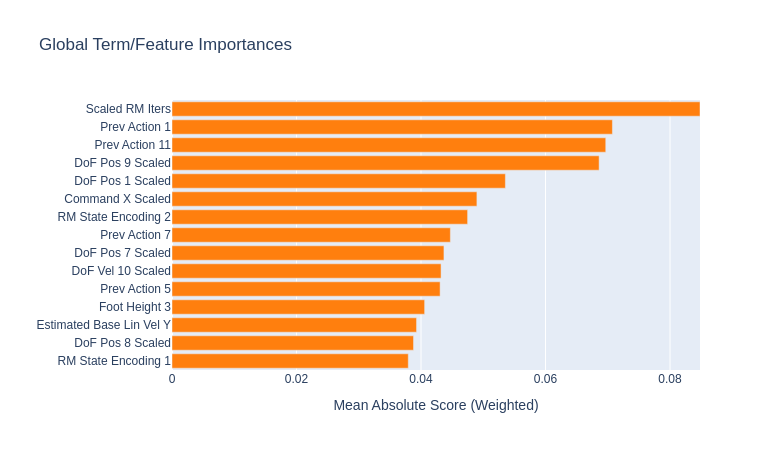}
    \includegraphics[width=1.09\columnwidth]{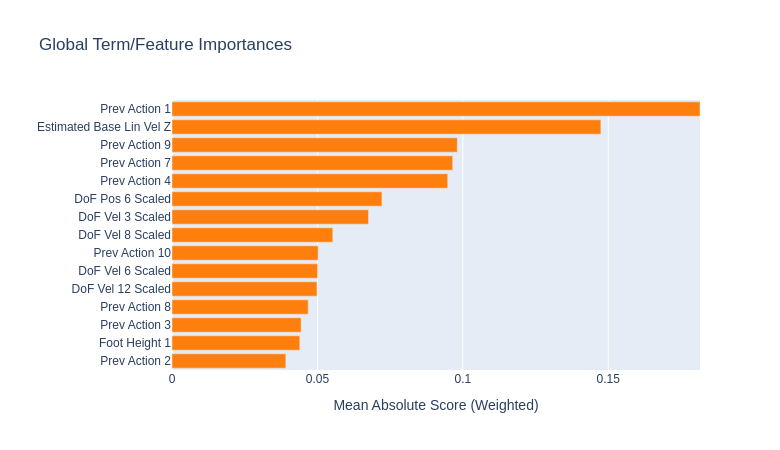}
    \caption{Global importances for EBM policies for walk, trot, pace, and bound (respectively, top to bottom) for Front Left Hip joint actions.}
    \label{fig:ebm-global-importances}
\end{figure}

Regarding EBMs, since they are modified GAMs, we can easily obtain the global importance of each term without additional models or computation, consisting of single and pairwise feature terms in Equation \ref{eq:EBM}. We provide the global importances of each term for distilled EBM policies in Figure \ref{fig:ebm-global-importances}. As detailed in \cite{nori2019interpretml}, EBMs are highly intelligible because the contribution of each term to the final prediction can be visualized and understood by plotting $f_i$ or $f_{i,j}$, providing global explanations of policy behaviour. 
Examples of such global explanations are presented in Figure \ref{fig:ebm-trot-feats}, which show the mapping learned by the policy for a specific observation or observation pair. We note how EBM top important features are different from GBM top important features, summarized in Table \ref{tab:top_features}, highlighting how observation importance for the same task varies depending on model architecture. 

Figure \ref{fig:ebm-trot-feats} shows how trot policy actions for the Front Left Hip joint are influenced by its foot height, as well as the pairwise interaction of the terms for previous action and joint position of the Hind Right Thigh. We note how this pairwise interaction map resembles a signed ``exclusive or'' operation, with near zero contribution to target joint angle in general, except for positive contributions when the previous action is positive and the joint position is negative, and negative contributions when the previous action is negative and the joint position is positive. EBMs also allow for local explanations, i.e. explaining the action corresponding to a specific input observation, as show in Figure \ref{fig:ebm-pace-local}. We argue this is particularly useful for safety certification or investigation purposes in the presence of malfunctions. Lastly, symbolic policies are interpretable in the sense that they constitute analytical expressions, and importance could be studied using partial derivatives, but we omit this due to their under-performance and for brevity (each policy has up to 90 terms). 





\section{CONCLUSION}

This work presents a novel approach for distilling neural network-based RL locomotion policies into interpretable ones, consisting of GBMs, EBMs, and symbolic policies for four gaits: walking, trotting, pacing, and bounding. Following the proposed methods, we conducted a thorough analysis of the performance and interpretability of distilled policies. Our results show that interpretable policies can be efficiently extracted from neural locomotion policies, which reveal valuable insights into the behaviour of RL locomotion policies and enable global and local explanations of the learned observation-action mapping, without compromising performance in the case of GBMs and EBMs. 

To the best of our knowledge, our work is the first to demonstrate that interpretable models can be used as policies for robot locomotion, and this work contributes to increased interpretability of RL locomotion policies. Future research directions include exploring the scalability of our approach to exteroceptive policies, incorporating uncertainty estimates, and extending our methodology to robot manipulation. We hope this work contributes towards enabling a widespread and trustworthy adoption of autonomous robots.

\bibliography{main}
\bibliographystyle{IEEEtran}

\end{document}